\documentclass{article} % For LaTeX2e
\usepackage{iclr2018_workshop,times}
\usepackage{hyperref}
\usepackage{url}
\usepackage{graphicx}
\usepackage{fixltx2e}
\usepackage{txfonts}
\usepackage{tabularx}
\usepackage{algorithmicx}
\usepackage{pgfplots}

\usepackage{algorithm}
\usepackage[noend]{algpseudocode}

\title{Neural Program Search: Solving Programming Tasks from Description and Examples}

\author{Illia Polosukhin \& Alex Skidanov \\
NEAR\\
\texttt{\{illia,alex\}@near.ai}
}

\begin{document}

\maketitle

\begin{abstract}
We present a \textsc{Neural Program Search}, an algorithm to generate programs from natural language description and a small number of input / output examples. The algorithm combines methods from Deep Learning and Program Synthesis fields by designing rich domain-specific language (DSL) and defining efficient search algorithm guided by a \textsc{Seq2Tree} model on it. To evaluate the quality of the approach we also present a semi-synthetic dataset of descriptions with test examples and corresponding programs. We show that our algorithm significantly outperforms a sequence-to-sequence model with attention baseline.
\end{abstract}

\section{Introduction}

The ability to synthesize a program from user intent (specification) is considered as one of the central problems in artificial intelligence (\citet{Green:1969:ATP:1624562.1624585}). Significant progress has been made recently in both program synthesis from examples (e.g. \citet{DBLP:journals/corr/BalogGBNT16}, \citet{polozov2015flashmeta}, \citet{learning-learn-programs-examples-going-beyond-program-structure}) and program synthesis from descriptions (e.g. \citet{program-synthesis-using-natural-language-2}, \citet{2017arXiv170900103Z}, \citet{LinWPVZE2017:TR}, \citet{DBLP:journals/corr/LingGHKSWB16}). 

Programming by example techniques such as Flash Fill (\citet{Gulwani2012SpreadsheetDM}) and BlinkFill (\citet{Singh2016BlinkFillSP}) were developed to help users perform data transformation tasks using examples instead of writing programs. These methods rely on a small domain-specific language (DSL) and then develop algorithms to efficiently search the space of programs. Two shortcomings of these approaches are that DSL limits types of programs that can be synthesized, and that large engineering effort is needed to fine-tune such systems.

Program synthesis from description has not been applied widely in practice yet. One of the challenges is that the natural language is very ambiguous, yet there are very strict requirements for the synthesized programs (see \citet{Yin2017ASN} and \citet{Rabinovich2017AbstractSN} for some discussion). In this paper we present \textsc{Neural Program Search} that learns from both description and examples and has high accuracy and speed to be applicable in practice.

We specifically consider a problem of synthesizing programs from a short description and several input / output pairs. By combining description and sample tests we address both limitations of programming by example and natural language program inference. We propose \texttt{LISP}-inspired DSL that is capable of representing solutions to many simple problems similar to those given as data transformation homework assignments, but is rather concise, making it more tractable to search in the space of programs in this DSL.

We propose a combination of two techniques --- search in the programs space that is guided by a deep learning model. This way we can use the latest advances in natural language understanding with the precision of the search techniques. We use a \textsc{Seq2Tree} model (\citet{alvarez2016tree}) that consists of a sequence encoder that reads the problem statement and a tree decoder augmented with attention that computes probabilities of each symbol in an AST tree node one node at a time. We then run a tree beam search that uses those probabilities to compute a number of most likely trees, and chooses one that is consistent with the given input/output pairs.

To evaluate the proposed model we have created a partially synthetic dataset \textsc{AlgoLISP} consisting of problem statements, solutions in our DSL and tests. We show that search guided by deep learning models achieves significantly better results than either of the two techniques separately.

\section{Related Work}

We describe the related work from domains of programming by example, programming from description, latent program induction and related field of semantic parsing.

\paragraph{Programming by Example}
 There were several practical applications of programming by example based on search techniques and carefully crafted heuristics, such as \citet{flashextract-framework-data-extraction-examples}. Also notable is recent work on application of deep learning for programming of examples such as RobustFill (\citet{Devlin2017RobustFillNP}) and combining deep learning models with traditional search techniques includes DeepCoder (\citet{DBLP:journals/corr/BalogGBNT16}), Neuro-Symbolic Program Synthesis (\citet{Parisotto2016NeuroSymbolicPS}) and Deep API Programmer (\citet{Bhupatiraju2017DeepAP}). "Neuro-Symbolic Program Synthesis" is similar to this work in that it predict tree structured programs and leverages that at search time. \citet{gaunt2016terpret} provides a comparison of various program synthesis from examples approaches on different benchmarks, showing limitations of existing gradient descent models.

\paragraph{Programming from Description}
Program synthesis from natural language descriptions has seen revival recently with progress in natural language understanding, examples of such work include \citet{program-synthesis-using-natural-language-2}, \citet{2017arXiv170900103Z}, \citet{LinWPVZE2017:TR} and \citet{DBLP:journals/corr/LingGHKSWB16}. Advances in this field are limited by small and/or noisy datasets and limitation of existing deep learning models when it comes to decoding highly structured sequences such as programs.

\paragraph{Latent Program Induction}
There has been a plethora of recent work in teaching neural networks the functional behavior of programs by augmenting the neural networks with additional computational modules such as Neural Turing Machines (\citet{Graves2014NeuralTM}), Neural GPUs (\citet{kaiser2015neural}), stacks-augmented RNNs (\citet{Joulin2015InferringAP}) and Neural Program-Interpreters (\citet{reed2015neural}). Two main limitations of these approaches are that these models must be trained separately for each task and that they do not expose interpretable program back to the user.

\paragraph{Semantic Parsing}
Semantic parsing is a related field to program synthesis from description, in which the space of programs is limited to some structured form. Noticeable work includes \citet{Dong2016LanguageTL} and \citet{Berant2013SemanticPO}. In another line of research latent programs for semantic parsing are learned from examples, e.g. \citet{neelakantan2016learning}, \citet{DBLP:journals/corr/LiangBLFL16}.
\section{Neural Program Search}

This section describes the DSL used for modeling, our neural network architecture 
and an algorithm for searching in program space.

\subsection{Domain Specific Language} \label{sec:dsl}

There are multiple reasons to use a domain specific language for code generation
instead of an existing programming language. One reason is to be able to convert a program to multiple target languages for practical applications (e.g. SQL, Python, Java),
which requires our DSL to be sufficiently general. Second, designing
a DSL from scratch allows to add constraints that would simplify its automated generation.

Our DSL is inspired by LISP -- functional language that can be easily represented
as an Abstract Syntax Tree and supports high-order functions. We augmented our DSL with a type system. While types do not appear in programs, each constant, argument or function has a type. A type is either an integer, a string, a boolean, a function or an array of other non-function types. 

A program in the DSL comprises a set of arguments (where each argument is defined by its name and type) and a program tree where each node belongs to one of the following symbol types: \textbf{constant}, \textbf{argument}, \textbf{function call}, \textbf{function}, or \textbf{lambda}. See Figure~\ref{fig:dsl} for a partial specification of the DSL.

The DSL also has a library of standard functions. Each function has a return type and a constant number of arguments, with each argument having its own type. The type system greatly reduces the number of possible combinations for each node in the program tree during search.

\begin{figure}[t!]
\centering
\begin{tabular}{lll}
\texttt{program} $\Coloneqq$ \texttt{symbol}\\
\texttt{symbol} $\Coloneqq$ \texttt{constant} $\vert$ \texttt{argument} $\vert$ \texttt{function\_call} $\vert$ \texttt{function} $\vert$ \texttt{lambda}\\
\texttt{constant} $\Coloneqq$ \texttt{number} $\vert$ \texttt{string} $\vert$ \textsc{True} $\vert$ \textsc{False}\\
\texttt{function\_call} $\Coloneqq$ (\texttt{function\_name} \texttt{arguments})\\
\texttt{function} $\Coloneqq$ \texttt{function\_name}\\
\texttt{arguments} $\Coloneqq$ \texttt{symbol} $\vert$ \texttt{arguments} \textsc{,} \texttt{symbol}\\
\texttt{function\_name} $\Coloneqq$ \textsc{reduce} $\vert$ \textsc{filter} $\vert$ \textsc{map} $\vert$ \textsc{head} $\vert$ \textsc{+} $\vert$ \textsc{-} \ldots \\
\texttt{lambda} $\Coloneqq$ \textsc{lambda} \texttt{function\_call} \\
\end{tabular}
\caption{Partial specification of the DSL used for this work.}
\label{fig:dsl}
\end{figure}

% As an example, consider the following program in the DSL that corresponds to the problem of finding the sum of all the fibonacci numbers from $1$ to $n$:
% 
% \begin{tabular}[c]{@{}l@{}}
% \texttt{(reduce,(range,1,n),0,}\\
% \texttt{~~(lambda1,}\\
% \texttt{~~~~(if,(<=,arg1,2),~~~~~~~~~~~~~~~~~~\# if the argument is <= 2}\\
% \texttt{~~~~~~~~1,~~~~~~~~~~~~~~~~~~~~~~~~~~~~\# the answer is 1}\\
% \texttt{~~~~~~~~(+,(self, (-, arg1, 1))~~~~~~~\# else it's a sum of}\\
% \texttt{~~~~~~~~~~~(self, (-, arg1, 2))))))~~~\#~~~two recursive calls to itself}\\
% \end{tabular}
% 
% Notice the use of $\texttt{arg1}$ and $\texttt{self}$ within the lambda.
%
%It might appear that this algorithm would invoke the lambda with the same parameters unnecessarily many times. However, due to the fact that functions in our DSL have no side effects, we can cache all the invocations, and thus this algorithm becomes linear.
%
\subsection{Seq2Tree}

Our neural network model uses an attentional encoder-decoder architecture.
The encoder uses RNN to embed concatenation of arguments \textsc{Args} and tokenized textual description of the task \textsc{Text}. The decoder is a doubly-recurrent neural network for generating tree structured output (\citet{alvarez2016tree}).
At each step of decoding, attention is used to augment current step with relevant information from encoder.

Formally, let $T = \{V, E, L\}$ be connected labeled tree, where $V$ is the set of nodes, $E$ is set of edges and $L$ are node labels. 
Let $H^e$ be a matrix of stacked problem statement encodings (outputs from encoder's RNN).
Let the $g^p$ and $g^s$ be functions which apply one step of the two separate RNNs.
For a node $i$ with parent $p(i)$ and previous sibling $s(i)$, the ancestral and fraternal hidden states are updated via:

\begin{equation} \label{hp_eqn}
    c^p_i = context(x_p(i), H^e) \quad\quad
    h^p_i = g^p(h^p_p(i), c^p_i)
\end{equation}
\begin{equation} \label{hs_eqn}
    c^s_i = context(x_s(i), H^e) \quad\quad
    h^s_i = g^s(h^s_s(i), c^s_i)
\end{equation}

where $x_p(i)$, $x_s(i)$ are the vectors representing the previous sibling’s and parent’s values, respectively.
And $context(x, H^e)$ computes current context using general attention mechanism (\citet{luong2015effective}) to align with encoder presentations using previous parent or sibling representation and 
combining it with $x$ in a non-linear way:

\begin{equation}
    a = softmax(H^e W_a x) \\
\end{equation}
\begin{equation}
    r = a^T H^e  \\
\end{equation}
\begin{equation}
    context =  tanh((r || x) W_c)
\end{equation}

where $||$ indicates vector concatenation and $W_a$ and $W_c$ are learnable parameters.
Once the hidden depth and width states have been updated with these observed labels, they are combined to obtain a full hidden state:

\begin{equation}
    h_i = (U^p h^p_i + U^s h^s_i)
\end{equation}

where $U_p$ and $U_s$ learnable parameters. This state contains combined information
from parent and siblings as well as attention to encoder representation and is used to predict label of the node.
In a simplest form (without placeholders), the label for node $i$ can be computed by sampling from distribution:

\begin{equation}
    o_i = softmax(Wh_i)
\end{equation}

After the node's output symbol $\hat{l}_i$ has been obtained by sampling from $o_i$, $x_i$ is obtained by embedding $\hat{l}_i$ using $W^T$.
Then the cell passes ($h^p_i$, $x_i$) to all it's children and ($h^s_i$, $x_i$) to the next sibling (if any), enabling them to apply Eqs (\ref{hp_eqn}) and (\ref{hs_eqn}) to compute their states.
This procedure continues recursively following schema defined by DSL that is being decoded.

The model is trained using back-propagation. The teacher forcing is used by using target topology of the code tree, feeding target labels for parent / sibling nodes.
The error is obtained using the cross-entropy loss of $o_i$ with respect to the true label $l_i$ for each decoded node.

Exploring alternative methods of training, such as \textsc{REINFORCE} (similar to \citet{2017arXiv170900103Z}) or using \textsc{Search} at training time is left for future work.

\begin{figure}[t!]
    \includegraphics[width=\textwidth]{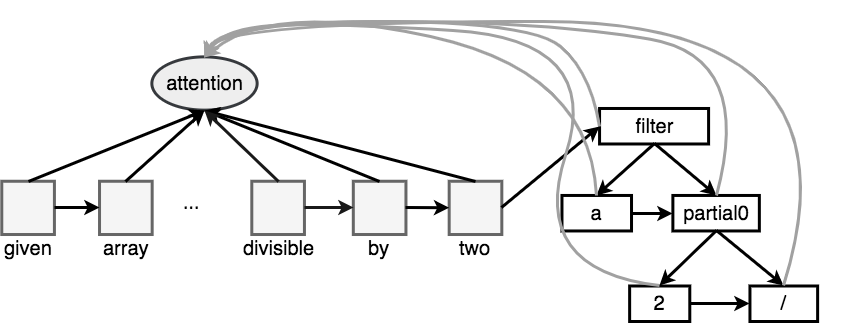}
    \caption{Example of \textsc{Seq2Tree} encoder-decoder model for "given an array, return values divisible by two". Left part is an encoder with embeddings+GRU cell, right is doubly-recurrent decoder with attention.}
    \label{fig:usage}
\end{figure}

\subsection{Search} \label{sec:search}

One of the central ideas of this work is to use Tree-Beam search in the program space using a deep learning model to score symbols in each AST node. The search continues until a complete program is found that passes given sample input / output pairs.

Search algorithm described in Algorithm~\ref{alg:tree-beam} starts with a priority queue with a single empty program.
At all times, we only keep the top \textsc{Queue\textsubscript{N}} most probable trees built so far in the priority queue.

\begin{algorithm}
\caption{Tree-Beam Search}
\label{alg:tree-beam}
\begin{algorithmic}[1]

    \State$queue \gets \texttt{HeapCreate}()$
    \State$model \gets \texttt{Seq2Tree}(task\_description)$
    \State$trees\_visited \gets 0$
    \State$\texttt{HeapPush}(queue, \texttt{EMPTY\_TREE})$
    \While{$\texttt{HeapLength}(queue) > 0 ~ \textbf{and} ~ trees\_visited < \texttt{MAX\_VISITED}$}
    \State$cur\_tree \gets \texttt{HeapPopMax}(queue)$
    \State$empty\_node \gets \texttt{FindFirstEmptyNode}(cur\_tree)$
    \If{$empty\_node = \texttt{null}$}
        \State$trees\_visited \gets trees\_visited + 1$
        \If{$\texttt{RunTests}(cur\_tree, sample\_tests) = \texttt{PASS}$}
            \Return{$cur\_tree$}
        \Else ~~ {$\textbf{continue}$}
        \EndIf
    \EndIf
    \ForAll{$(prob, symbol)$ in $\texttt{GetProbs}(model, empty\_node)$} \Comment{In decreasing order of probabilities}
        \If{$prob < \texttt{THRESHOLD}$} ~~ $\textbf{break}$
        \EndIf

        \If{$\texttt{SymbolMatchesNodeType}(symbol, empty\_node)$}
            \State$new\_tree \gets \texttt{CloneTreeAndSubstitute}(cur\_tree, empty\_node, symbol)$
            \State$\texttt{HeapPush}(queue, new\_tree)$
        \EndIf

    \EndFor
    \While{$\texttt{HeapLength}(queue) > \texttt{QUEUE\_N}$}
        \State$\texttt{HeapPopMin}(queue)$
    \EndWhile
    \EndWhile
    \State \Return $\texttt{null}$

\hspace*{-19mm}
\end{algorithmic}
\end{algorithm}

If a program on the top of the queue is complete (no more nodes need to be added), 
we run evaluation with given sample input / output examples. 
If the results from current program match expected outputs, search is stopped.
Alternatively, if over \textsc{MAX\_VISITED} programs have already been evaluated, 
the search stops without program found.

Each program in the priority queue is represented as an incomplete tree with some nodes already synthesized and some still empty.
When such incomplete tree $T$ is popped from the queue, we locate the first empty node $n$ in the pre-order traversal of the tree, 
and use \textsc{Seq2Tree} model to compute probabilities of each possible symbol being in that node. At that point we already know
the type of the symbol the node should contain, and thus only consider symbols of that type. For each such symbol $s$ we construct
a new tree by replacing $n$ with $s$. We then
push all the new trees, no matter how unlikely they are into the priority queue, and then remove least probable trees until the
size of the priority queue is not \textsc{Queue\textsubscript{N}} or less.

\begin{figure}[t!]
    \includegraphics[width=\textwidth]{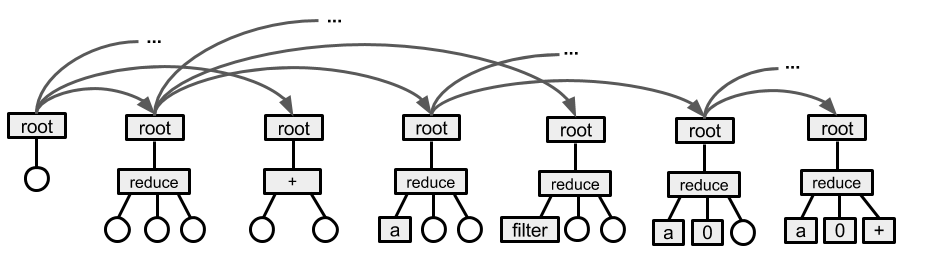}
    \caption{Example of tree search for a query "Given an array, find the sum of its elements". Rectangles represent nodes with a symbol, while circles represent empty nodes. We start with an empty tree on the far left. When that tree is popped from the priority queue, we consider each possible symbol for the first empty node in the pre-order traversal, and create a new tree for each. Two such trees are shown in this figure, for symbols \textit{reduce} and \textit{+}. When the tree with \textit{reduce} is popped, several new trees are generated by filling in the first empty node in the pre-order traversal of that tree, which is the first child of reduce. The first argument of reduce is an array, so only symbols that produce arrays are considered. Two trees for such symbols are shown on the figure for \textit{a}, which is an argument, and \textit{filter}. The search continues until either \textsc{D} trees are generated, or a tree that passes all the sample tests is found. Such tree is shown on the far right.}
    \label{fig:usage}
\end{figure}

In our experiments evaluating the \textsc{Seq2Tree} model takes comparable amount of time to cloning trees and pushing them to the queue, so optimizing both steps would contribute to the performance of the search. We use the following optimization techniques:

\textbf{Persistent trees}. After evaluating the model once we need to clone the tree as many times as many symbols will be considered for the first empty node. Storing all the trees in full can be memory consuming, and, depending on the language in which the search is implemented, allocating objects for the nodes can take considerable amount of time. One way to save memory and time on cloning trees is to use persistent trees. When a new tree $T_{new}$ is created from a tree $T$ by introducing a new node $s$, it is sufficient to clone only the nodes on the path from root to $s$, and replace their corresponding children on that path with the cloned version. This takes time and memory proportional to the height of the tree, which for larger trees is significantly smaller than the total number of nodes in the tree. The tree is then represented as a pointer to the root node.

\textbf{Batched search}. During training we need to read trees of different shapes, which is a challenging problem, and we use dynamic batching to address it. During search we only invoke \textsc{Seq2Tree} on a single node, so multiple such invocations can be trivially batched. We batch invocations of the \textsc{Seq2Tree} across tasks in the following way: we run the search for $batch\_size$ tasks simultaneously, and on each step pop the single most likely incomplete tree for each task, identify the empty node in each of them, and compute the probabilities of the symbols in all of them at once.

This approach speeds up evaluation of the search if it is run on multiple tasks simultaneously, for example when evaluating the accuracy on a held out set. However, in cases when only one task is being evaluated batching across tasks is not applicable. We evaluated the following alternative: on each iteration of search pop the top $batch\_size$ incomplete trees from the priority queue instead of just the single most likely one, identify the empty node in each of them, and compute the probabilities of symbols in all of them at once. This approach did not produce any noticeable speed up, and in most cases even slowed the search down slightly. The possible reason for that is that if the model that guides the search is good, the correct incomplete tree will be the top one most of the time, so number of model evaluations that are saved due to batched execution is very small, and the extra computation time of evaluating a model on a batch instead of a single sample outweighs the time saved due to those few extra evaluations.

\section{AlgoLisp} \label{sec:dataset}

In this section we describe a new dataset we prepared to train and evaluate models that learn
to synthesize simple data processing programs.

\textsc{AlgoLisp} is a dataset of problem descriptions, corresponding implementations
of the problem in a \textsc{Lisp}-inspired programming language described in section \ref{sec:dsl} and tests.
Each problem has 10 tests, where each test is input to be fed into the synthesized program and 
the expected output the program should produce. All the problems are designed 
in such way that the output for each input is unique.  See Table~\ref{data-stats} for dataset statistics.

There are multiple existing datasets for code synthesis task from natural language. 
Some recent notable ones are description to bash command (\citet{LinWPVZE2017:TR}) and description to SQL 
(\citet{2017arXiv170900103Z}). To the best of our knowledge, no existing dataset is applicable
to our problem due to reasons such as no easy way of evaluating the results, insufficient
complexity of the programs or size too small for deep learning.

{
\renewcommand{\arraystretch}{1.2}

\begin{table}[t]
    \caption{\textsc{AlgoLisp} statistics.}
    \label{data-stats}
    \begin{center}
    \begin{tabular}{llll}
        \hline
       &\multicolumn{1}{c}{\bf Train}  &\multicolumn{1}{c}{\bf Dev} &\multicolumn{1}{c}{\bf Test}
    \\ \hline
    \# tasks         &$79,214$ &$9,352$  &$10,940$ \\
    Avg text len     &38.17    &39.95    &37.58   \\
    Avg code depth   &7.61     &8.23     &7.97    \\
    Avg code len     &24.33    &29.31    &27.16   \\
    Vocab size      &\multicolumn{3}{c}{230}      \\ \hline
    \end{tabular}
    \end{center}
\end{table}
}
    
Because same problem can be solved with many different programs, the solution is considered correct if it produces correct output on 
all the tests for given problem. For consistency and comparable results we suggest two specific 
approaches in which the tests are used during inference: no tests using at inference time (used for deep learning only models) and using first 3 tests for search and the remaining 7 tests as a holdout to evaluate correctness of the found program.

The dataset was synthesized with the following procedure (see Table~\ref{dataset-gen} for examples). We first chose several dozen tasks from homework 
assignments for basic computer science and algorithms courses. For each task, we parameterized assignments (e.g. in statement "find all even elements in an array" \textit{even} could be replaced by \{prime, even, odd, divisible by three, positive, negative\})and matching code. The final dataset is then random combination of such tasks, where other tasks can be passed into the given statement as input (e.g. two statements "find all even elements in an array" and "sort an array" will be combined to "find all even elements in an array and return them in sorted order").

This dataset is designed for the task of learning basic composition and 
learning to use simple concepts and routines in the \textsc{DSL}. Due to the fact that the 
number of homework assignments used for this dataset was relatively low, it is unlikely 
that the models trained on this dataset would generalize to new types of algorithm.

To make sure that the models are learning to compose simpler concepts for novel problems, the dataset split into train, dev, and test by surface form of the code. Thus ensuring that at training time the model has not observed any programs it will be evaluated on.

To evaluate neural networks and search driven algorithms, we compare output of the generated programs on a holdout set of tests for each task. Thus accuracy on this dataset is defined as \textsc{Acc} = \(\frac{N\textsubscript{C}}{N}\), where \textsc{N} is total number of tasks and \textsc{N\textsubscript{C}} is number of tasks for which the synthesized solution passes all the holdout tests.

\begin{table}[t]
    \caption{
        Examples from \textsc{AlgoLISP} dataset. First row is an example of user provided homework assignment with program in our DSL. Subsequent lines are examples of synthesized tasks and programs, showing various properties of the generator: different text for the task, combination with other sub-problems (such as "elements in $a$ that are present in $b$") and variation of task properties.}
    \label{dataset-gen}
    %\begin{center}
    \begin{tabular}{|l|l|}
        \hline
        \begin{tabular}[c]{@{}l@{}}
            You are given an array $a$. Find the smallest, \\ 
            element in $a$, which is strictly greater than \\
            the minimum element in $a$.
        \end{tabular}
        &
        \begin{tabular}[c]{@{}l@{}}
            \texttt{(reduce (filter a }\\
            \texttt{~~(partial0 (reduce a inf) <))}\\
            \texttt{~~inf min)}
        \end{tabular}
        \\ \hline

        \multicolumn{2}{|c|}{\textbf{Synthesized examples}} \\ \hline

        \begin{tabular}[c]{@{}l@{}}
            Consider an array of numbers $a$, \\ 
            your task is to compute largest element among \\
            values in $a$, which is strictly smaller than  \\
            the maximum element among values in $a$.
        \end{tabular}
        &
        \begin{tabular}[c]{@{}l@{}}
            \texttt{(reduce (filter a }\\
            \texttt{~~(partial0 (reduce a -inf) >))}\\
            \texttt{~~-inf max)}
        \end{tabular}
        \\ \hline
        
        \begin{tabular}[c]{@{}l@{}}
            Given arrays of numbers $a$ and $b$, compute \\
            largest element among elements in $a$ that are\\
            present in $b$, which is strictly less than \\
            maximum element among elements in $a$ that \\
            are present in $b$.
        \end{tabular}
        & 
        \begin{tabular}[c]{@{}l@{}}
            \texttt{(reduce (filter }\\
            \texttt{~~(filter a (partial0 b contains))}\\
            \texttt{~~(partial0 (reduce }\\
            \texttt{~~~~(filter a (partial0 b contains)) }\\
            \texttt{~~~~inf) <))}\\
            \texttt{~~inf min)}
        \end{tabular}
        \\ \hline
        
        \begin{tabular}[c]{@{}l@{}}
            Given an array of numbers, your task is to \\
            find largest element among values in the \\
            given array that are divisible by two, \\
            which is strictly less than maximum \\
            element among values in the given \\
            array that are divisible by two.
        \end{tabular}
        & 
        \begin{tabular}[c]{@{}l@{}}
            \texttt{(reduce (filter }\\
            \texttt{~~(filter a is\_odd) }\\
            \texttt{~~(partial0 (reduce }\\
            \texttt{~~~~(filter a is\_odd)}\\
            \texttt{~~~~-inf) >))}\\
            \texttt{~~-inf max)}
        \end{tabular}
        \\ \hline
    \end{tabular}
    %\end{center}
\end{table}

\section{Experiments}

We implemented all models using PyTorch\footnote{http://pytorch.org} and used 
Dynamic Batching\footnote{We used implementation described in http://near.ai/articles/2017-09-06-PyTorch-Dynamic-Batching/} (e.g. \citet{neubig2017fly}) to implement batched tree decoding at training time. 
We train using ADAM (\citet{kingma2014adam}), embedding and recurrent layers have hidden size of 100 units.

The placeholders are used to handle OOV (\citet{hewlett2016wikireading}) in all neural networks. Placeholders are added to the vocabulary, increasing the vocabulary size from $N_v$ to $N_v + N_p$, where $N_p$ is a fixed size number of placeholders, selected to be larger than number of tokens in the input.
The same OOV tokens from inputs and outputs are mapped to the same placeholder (selected at random from not used yet), allowing model to attend and generate them at decoding time. Given the attention mechanism this is very similar to Pointer Networks (\citet{Vinyals2015PointerN}).

\subsection{Results}

We compare our model with Attentional Sequence to Sequence similar to \citet{luong2015effective}. Sequence to sequence models have shown near state of the art results at machine translation, question answering and semantic parsing. Additional baseline we compared to model that synthesizes program from examples - \textsc{IO2Seq}, inspired by RobustFill \citet{Devlin2017RobustFillNP}. The model reads all inputs and output via byte encoder for each given test case, max-pools their encoding into single vector and does sequential decoding of program.

The Table~\ref{result-synth} presents results on \textsc{AlgoLisp} dataset
for \textsc{Seq2Seq+Att}, \textsc{IO2Sec} and \textsc{Seq2Tree} model with and without applying search described in section \ref{sec:search}. Additionally performance of the \textsc{Search} on its own is presented, to show result of search through program space without machine learning model guidance by only validating on input / output examples.

Explicitly modeling tree structure of code in \textsc{Seq2Tree} improves upon attentional sequence to sequence model by $11\%$. \textsc{Search} on it's own finds very limited number of programs with the same limit $MAX\_VISITED = 100$ (see \ref{sec:analysis} for details) as \textsc{Seq2Tree + Search}. Only IO model 
\textsc{IO2Seq} also doesn't perform particularly well even augmented with search, as there is a lot of problems where just test cases are not enough to recover the underlying program. Final model \textsc{Seq2Tree + Search} combines both neural and search into one model and improves to the best result -- $85.8\%$.

{
\renewcommand{\arraystretch}{1.2}

\begin{table}[t]
\caption{Performance on \textsc{AlgoLisp}. Accuracy is defined in section \ref{sec:dataset}.}
\label{result-synth}
\begin{center}
\begin{tabular}{lll}
\hline
    \multicolumn{1}{c}{\bf Model}  
    & \multicolumn{1}{c}{\bf Dev Acc}
    & \multicolumn{1}{c}{\bf Test Acc}
\\ \hline 
Attentional Seq2Seq         &$54.4\%$           &$54.1\%$ \\
Attentional Seq2Seq + Search &$72.6\%$           &$72.3\%$ \\ \hline
IO2Seq                      &$2.5\%$             &$2.2\%$ \\
IO2Seq + Search             &$13.3\%$            &$12.8\%$ \\ \hline
Search                      & $0.5\%$           & $0.6\%$ \\
Seq2Tree                    &$61.2\%$           &$61.0\%$ \\
\textbf{Seq2Tree + Search}  &$\textbf{86.1\%}$  &$\textbf{85.8\%}$ \\ \hline
\end{tabular}
\end{center}
\end{table}

}

\subsection{Analysis} \label{sec:analysis}

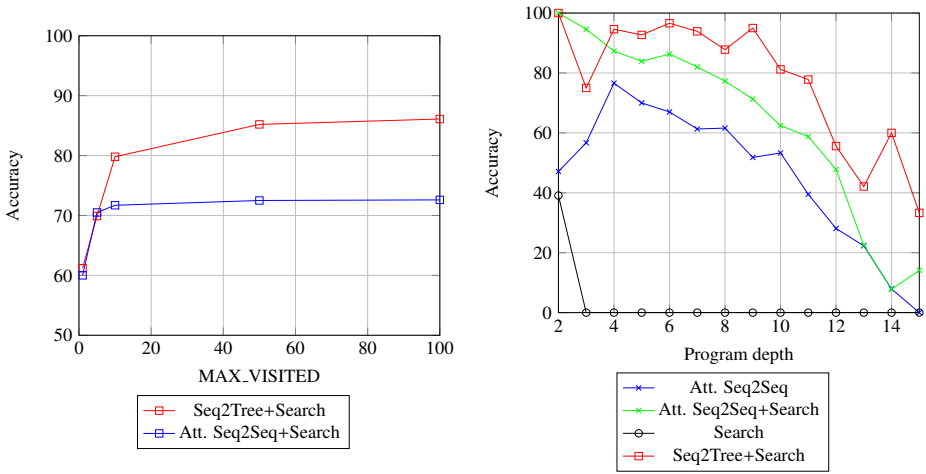
\begin{figure}[h]
    \centering
    \begin{minipage}{.45\textwidth}
    \begin{tikzpicture}[scale=0.7]
        % \pdfplotsset{
        %     xmajorgrids=true,
        %     ymajorgrids=true,
        % }
        % \begin{axis}[
        %         axis y line*=right,
        %         xmin=0, xmax=100, ymin=0, ymax=1,
        %         legend style={at={(0.75,-0.2)}},
        %         xlabel={MAX\_VISITED},
        %         ylabel={Accuracy},
        %     ]        
        %     \addplot[color=blue,mark=square]
        %     coordinates {
        %         (1, 0.0)(5, 0.0)(10, 0.0)(15, 0.1)(20, 0.1)(25, 0.1)(30, 0.1)(35, 0.1)(40, 0.2)(45, 0.3)(50, 0.3)(55, 0.4)(65, 0.6)(70, 0.6)(75, 0.6)(80, 0.6)(85, 0.6)(90, 0.6)(95, 0.6)(100, 0.6)
        %     };
        % \end{axis}
        \begin{axis}[
                xmajorgrids=true,
                ymajorgrids=true,
                xmin=0, xmax=100, ymin=50, ymax=100,
                legend style={at={(0.75,-0.2)}},
                xlabel={MAX\_VISITED},
                ylabel={Accuracy},
            ]        
            \addplot[color=red,mark=square]
                coordinates {
                    (1, 61.2)(5, 69.9)(10, 79.8)(50, 85.2)(100, 86.1)
                };
            \addplot[color=blue,mark=square]
                coordinates {
                    (1, 60.0)(5, 70.5)(10, 71.7)(50, 72.5)(100, 72.6)
                };
            \legend{Seq2Tree+Search,Att. Seq2Seq+Search}
        \end{axis}
    \end{tikzpicture}
    \end{minipage}
    \begin{minipage}{.45\textwidth}
    \begin{tikzpicture}[scale=0.7]
            \begin{axis}[
                xmajorgrids=true,
                ymajorgrids=true,
                xmin=2, xmax=15, ymin=0, ymax=100,
                legend style={at={(0.75,-0.2)}},
                xlabel={Program depth},
                ylabel={Accuracy},
            ]
                \addplot[
                    color=blue,
                    mark=x,
                    ]
                    coordinates {
                        (2, 47.1)(3, 56.7)(4, 76.6)(5, 70.0)(6, 67.0)(7, 61.3)(8, 61.6)(9, 51.8)(10, 53.3)(11, 39.5)(12, 28.1)(13, 22.3)(14, 7.9)(15, 0.0)
                    };
                \addplot[
                    color=green,
                    mark=x,
                    ]
                    coordinates {
                        (2, 100.0)(3, 94.6)(4, 87.3)(5, 83.9)(6, 86.3)(7, 82.0)(8, 77.3)(9, 71.3)(10, 62.4)(11, 58.8)(12, 47.8)(13, 22.5)(14, 7.8)(15, 14.1)
                    };
                \addplot[
                    color=black,
                    mark=o,
                    ]
                    coordinates {
                        (2, 39.1)(3, 0.0)(4, 0.0)(5, 0.0)(6, 0.0)(7, 0.0)(8, 0.0)(9, 0.0)(10, 0.0)(11, 0.0)(12, 0.0)(13, 0.0)(14, 0.0)(15, 0.0)
                    };
                \addplot[
                    color=red,
                    mark=square,
                    ]
                    coordinates {
                        (2, 100.0)(3, 75.0)(4, 94.6)(5, 92.7)(6, 96.6)(7, 93.9)(8, 87.8)(9, 95.0)(10, 81.2)(11, 77.8)(12, 55.6)(13, 42.1)(14, 60.0)(15, 33.3)
                    };
                \legend{Att. Seq2Seq,Att. Seq2Seq+Search,Search,Seq2Tree+Search}
            \end{axis}
        \end{tikzpicture}
    \end{minipage}
    
        \caption{Analysis of results on dev set. Left plot shows accuracy of the model varying \textsc{MAX\_VISITED} in \textsc{Search} algorithm. Right plot shows accuracy stratified by depth of the target code tree.}
        \label{fig:fig-analysis}
\end{figure}

\textbf{The pattern of how accuracy changes with the number of trees visited during search shows the quality of the neural network}. In general, given no limit on $MAX\_VISITED$, \textsc{Search} will explore the entirety of the program space and find all programs that solve sample tests, which in our case contains on the order of $10^{2^D}$ programs, where $D$ is depth of programs explored. To compare improvement that neural network model brings to search, we compare the model performance at different thresholds of $MAX\_VISITED$. See Figure~\ref{fig:fig-analysis} for results. As expected, the accuracy of the model grows if the search gets to explore more trees. Interestingly, the growth of accuracy of \textsc{Seq2Tree+Search} slows down very quickly. It is expected if the neural network is good, since then it predicts correct symbols with high accuracy, and therefore the correct tree is more likely to be found early during the search.

\textbf{Depth of the program is a reasonable proxy for complexity of the problem}. Right part of Figure~\ref{fig:fig-analysis} shows accuracy of the models based on gold program depth. Note that there are relatively few programs with depth below $5$ in the dev set, which leads to higher variance. As expected, with the growth of the depth of the tree, the accuracy reduces, since more nodes need to be predicted.

\section{Conclusion}

We have presented an algorithm for program synthesis from textual specification and a sample of input / output pairs, 
that combines deep learning network for understanding language and general programming patterns with
conventional search technique that allows to find correct program in discrete space which neural models struggle with. We presented a semi-synthetic dataset to empirically evaluate learning of program composition and usage of programing constructs. Our empirical results show improvement using combination of structured tree decoding and search over attentional sequence to sequence model.

There remain some limitations, however. 
Our training data currently is semi-generated and contains only limited set of types of problems. It is prohibitively expensive to collect a human annotated set with large quantity of tasks per each problem type, so finding a way to learn from few examples per problem type is crucial. Additionally, in many practical use cases there will be no input / output examples, requiring interaction with the user to resolve ambiguity and improved techniques for structural output decoding in neural networks.

\bibliography{iclr2018_workshop}

\begin{thebibliography}{31}
\providecommand{\natexlab}[1]{#1}
\providecommand{\url}[1]{\texttt{#1}}
\expandafter\ifx\csname urlstyle\endcsname\relax
  \providecommand{\doi}[1]{doi: #1}\else
  \providecommand{\doi}{doi: \begingroup \urlstyle{rm}\Url}\fi

\bibitem[Alvarez-Melis \& Jaakkola(2016)Alvarez-Melis and
  Jaakkola]{alvarez2016tree}
David Alvarez-Melis and Tommi~S Jaakkola.
\newblock Tree-structured decoding with doubly-recurrent neural networks.
\newblock 2016.

\bibitem[Balog et~al.(2016)Balog, Gaunt, Brockschmidt, Nowozin, and
  Tarlow]{DBLP:journals/corr/BalogGBNT16}
Matej Balog, Alexander~L. Gaunt, Marc Brockschmidt, Sebastian Nowozin, and
  Daniel Tarlow.
\newblock Deepcoder: Learning to write programs.
\newblock \emph{CoRR}, abs/1611.01989, 2016.
\newblock URL \url{http://arxiv.org/abs/1611.01989}.

\bibitem[Berant et~al.(2013)Berant, Chou, Frostig, and
  Liang]{Berant2013SemanticPO}
Jonathan Berant, Andrew Chou, Roy Frostig, and Percy Liang.
\newblock Semantic parsing on freebase from question-answer pairs.
\newblock In \emph{EMNLP}, 2013.

\bibitem[Bhupatiraju et~al.(2017)Bhupatiraju, Singh, rahman Mohamed, and
  Kohli]{Bhupatiraju2017DeepAP}
Surya Bhupatiraju, Rishabh Singh, Abdel rahman Mohamed, and Pushmeet Kohli.
\newblock Deep api programmer: Learning to program with apis.
\newblock \emph{CoRR}, abs/1704.04327, 2017.

\bibitem[Desai et~al.(2016)Desai, Gulwani, Hingorani, Jain, Karkare, Marron, R,
  and Roy]{program-synthesis-using-natural-language-2}
Aditya Desai, Sumit Gulwani, Vineet Hingorani, Nidhi Jain, Amey Karkare, Mark
  Marron, Sailesh R, and Subhajit Roy.
\newblock Program synthesis using natural language.
\newblock May 2016.
\newblock URL
  \url{https://www.microsoft.com/en-us/research/publication/program-synthesis-using-natural-language-2/}.

\bibitem[Devlin et~al.(2017)Devlin, Uesato, Bhupatiraju, Singh, rahman Mohamed,
  and Kohli]{Devlin2017RobustFillNP}
Jacob Devlin, Jonathan Uesato, Surya Bhupatiraju, Rishabh Singh, Abdel rahman
  Mohamed, and Pushmeet Kohli.
\newblock Robustfill: Neural program learning under noisy i/o.
\newblock In \emph{ICML}, 2017.

\bibitem[Dong \& Lapata(2016)Dong and Lapata]{Dong2016LanguageTL}
Li~Dong and Mirella Lapata.
\newblock Language to logical form with neural attention.
\newblock \emph{CoRR}, abs/1601.01280, 2016.

\bibitem[Ellis \& Gulwani(2017)Ellis and
  Gulwani]{learning-learn-programs-examples-going-beyond-program-structure}
Kevin Ellis and Sumit Gulwani.
\newblock Learning to learn programs from examples: Going beyond program
  structure.
\newblock May 2017.
\newblock URL
  \url{https://www.microsoft.com/en-us/research/publication/learning-learn-programs-examples-going-beyond-program-structure/}.

\bibitem[Gaunt et~al.(2016)Gaunt, Brockschmidt, Singh, Kushman, Kohli, Taylor,
  and Tarlow]{gaunt2016terpret}
Alexander~L Gaunt, Marc Brockschmidt, Rishabh Singh, Nate Kushman, Pushmeet
  Kohli, Jonathan Taylor, and Daniel Tarlow.
\newblock Terpret: A probabilistic programming language for program induction.
\newblock \emph{arXiv preprint arXiv:1608.04428}, 2016.

\bibitem[Graves et~al.(2014)Graves, Wayne, and Danihelka]{Graves2014NeuralTM}
Alex Graves, Greg Wayne, and Ivo Danihelka.
\newblock Neural turing machines.
\newblock \emph{CoRR}, abs/1410.5401, 2014.

\bibitem[Green(1969)]{Green:1969:ATP:1624562.1624585}
Cordell Green.
\newblock Application of theorem proving to problem solving.
\newblock In \emph{Proceedings of the 1st International Joint Conference on
  Artificial Intelligence}, IJCAI'69, pp.\  219--239, San Francisco, CA, USA,
  1969. Morgan Kaufmann Publishers Inc.
\newblock URL \url{http://dl.acm.org/citation.cfm?id=1624562.1624585}.

\bibitem[Gulwani(2014)]{flashextract-framework-data-extraction-examples}
Sumit Gulwani.
\newblock Flashextract: A framework for data extraction by examples.
\newblock June 2014.
\newblock URL
  \url{https://www.microsoft.com/en-us/research/publication/flashextract-framework-data-extraction-examples/}.

\bibitem[Gulwani et~al.(2012)Gulwani, Harris, and
  Singh]{Gulwani2012SpreadsheetDM}
Sumit Gulwani, William~R. Harris, and Rishabh Singh.
\newblock Spreadsheet data manipulation using examples.
\newblock \emph{Commun. ACM}, 55:\penalty0 97--105, 2012.

\bibitem[Hewlett et~al.(2016)Hewlett, Lacoste, Jones, Polosukhin, Fandrianto,
  Han, Kelcey, and Berthelot]{hewlett2016wikireading}
Daniel Hewlett, Alexandre Lacoste, Llion Jones, Illia Polosukhin, Andrew
  Fandrianto, Jay Han, Matthew Kelcey, and David Berthelot.
\newblock Wikireading: A novel large-scale language understanding task over
  wikipedia.
\newblock \emph{arXiv preprint arXiv:1608.03542}, 2016.

\bibitem[Joulin \& Mikolov(2015)Joulin and Mikolov]{Joulin2015InferringAP}
Armand Joulin and Tomas Mikolov.
\newblock Inferring algorithmic patterns with stack-augmented recurrent nets.
\newblock In \emph{NIPS}, 2015.

\bibitem[Kaiser \& Sutskever(2015)Kaiser and Sutskever]{kaiser2015neural}
{\L}ukasz Kaiser and Ilya Sutskever.
\newblock Neural gpus learn algorithms.
\newblock \emph{arXiv preprint arXiv:1511.08228}, 2015.

\bibitem[Kingma \& Ba(2014)Kingma and Ba]{kingma2014adam}
Diederik Kingma and Jimmy Ba.
\newblock Adam: A method for stochastic optimization.
\newblock \emph{arXiv preprint arXiv:1412.6980}, 2014.

\bibitem[Liang et~al.(2016)Liang, Berant, Le, Forbus, and
  Lao]{DBLP:journals/corr/LiangBLFL16}
Chen Liang, Jonathan Berant, Quoc Le, Kenneth~D. Forbus, and Ni~Lao.
\newblock Neural symbolic machines: Learning semantic parsers on freebase with
  weak supervision.
\newblock \emph{CoRR}, abs/1611.00020, 2016.
\newblock URL \url{http://arxiv.org/abs/1611.00020}.

\bibitem[Lin et~al.(2017)Lin, Wang, Pang, Vu, Zettlemoyer, and
  Ernst]{LinWPVZE2017:TR}
Xi~Victoria Lin, Chenglong Wang, Deric Pang, Kevin Vu, Luke Zettlemoyer, and
  Michael~D. Ernst.
\newblock Program synthesis from natural language using recurrent neural
  networks.
\newblock Technical Report UW-CSE-17-03-01, University of Washington Department
  of Computer Science and Engineering, Seattle, WA, USA, March 2017.

\bibitem[Ling et~al.(2016)Ling, Grefenstette, Hermann, Kocisk{\'{y}}, Senior,
  Wang, and Blunsom]{DBLP:journals/corr/LingGHKSWB16}
Wang Ling, Edward Grefenstette, Karl~Moritz Hermann, Tom{\'{a}}s Kocisk{\'{y}},
  Andrew Senior, Fumin Wang, and Phil Blunsom.
\newblock Latent predictor networks for code generation.
\newblock \emph{CoRR}, abs/1603.06744, 2016.
\newblock URL \url{http://arxiv.org/abs/1603.06744}.

\bibitem[Luong et~al.(2015)Luong, Pham, and Manning]{luong2015effective}
Minh-Thang Luong, Hieu Pham, and Christopher~D Manning.
\newblock Effective approaches to attention-based neural machine translation.
\newblock \emph{arXiv preprint arXiv:1508.04025}, 2015.

\bibitem[Neelakantan et~al.(2016)Neelakantan, Le, Abadi, McCallum, and
  Amodei]{neelakantan2016learning}
Arvind Neelakantan, Quoc~V Le, Martin Abadi, Andrew McCallum, and Dario Amodei.
\newblock Learning a natural language interface with neural programmer.
\newblock \emph{arXiv preprint arXiv:1611.08945}, 2016.

\bibitem[Neubig et~al.(2017)Neubig, Goldberg, and Dyer]{neubig2017fly}
Graham Neubig, Yoav Goldberg, and Chris Dyer.
\newblock On-the-fly operation batching in dynamic computation graphs.
\newblock \emph{arXiv preprint arXiv:1705.07860}, 2017.

\bibitem[Parisotto et~al.(2016)Parisotto, rahman Mohamed, Singh, Li, Zhou, and
  Kohli]{Parisotto2016NeuroSymbolicPS}
Emilio Parisotto, Abdel rahman Mohamed, Rishabh Singh, Lihong Li, Dengyong
  Zhou, and Pushmeet Kohli.
\newblock Neuro-symbolic program synthesis.
\newblock \emph{CoRR}, abs/1611.01855, 2016.

\bibitem[Polozov \& Gulwani(2015)Polozov and Gulwani]{polozov2015flashmeta}
Oleksandr Polozov and Sumit Gulwani.
\newblock Flashmeta: A framework for inductive program synthesis.
\newblock \emph{ACM SIGPLAN Notices}, 50\penalty0 (10):\penalty0 107--126,
  2015.

\bibitem[Rabinovich et~al.(2017)Rabinovich, Stern, and
  Klein]{Rabinovich2017AbstractSN}
Maxim Rabinovich, Mitchell Stern, and Dan Klein.
\newblock Abstract syntax networks for code generation and semantic parsing.
\newblock In \emph{ACL}, 2017.

\bibitem[Reed \& De~Freitas(2015)Reed and De~Freitas]{reed2015neural}
Scott Reed and Nando De~Freitas.
\newblock Neural programmer-interpreters.
\newblock \emph{arXiv preprint arXiv:1511.06279}, 2015.

\bibitem[Singh(2016)]{Singh2016BlinkFillSP}
Rishabh Singh.
\newblock Blinkfill: Semi-supervised programming by example for syntactic
  string transformations.
\newblock \emph{PVLDB}, 9:\penalty0 816--827, 2016.

\bibitem[Vinyals et~al.(2015)Vinyals, Fortunato, and
  Jaitly]{Vinyals2015PointerN}
Oriol Vinyals, Meire Fortunato, and Navdeep Jaitly.
\newblock Pointer networks.
\newblock In \emph{NIPS}, 2015.

\bibitem[Yin \& Neubig(2017)Yin and Neubig]{Yin2017ASN}
Pengcheng Yin and Graham Neubig.
\newblock A syntactic neural model for general-purpose code generation.
\newblock In \emph{ACL}, 2017.

\bibitem[{Zhong} et~al.(2017){Zhong}, {Xiong}, and
  {Socher}]{2017arXiv170900103Z}
V.~{Zhong}, C.~{Xiong}, and R.~{Socher}.
\newblock {Seq2SQL: Generating Structured Queries from Natural Language using
  Reinforcement Learning}.
\newblock \emph{ArXiv e-prints}, August 2017.

\end{thebibliography}
\bibliographystyle{iclr2018_workshop}

\end{document}